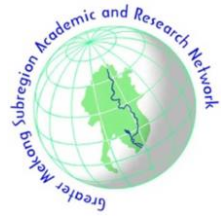

# Validation of a 24-hour-ahead Prediction model for a Residential Electrical Load under diverse climate

Ehtisham Asghar[1], Martin Hill[1], Ibrahim Sengor[1], Conor Lynch[2], and
Phan Quang An[3, *]

**A B S T R A C T**

Accurate household electrical energy demand prediction is essential for effectively managing sustainable Energy Communities. Integrated with the Energy Management System, these communities aim to optimise operational costs. However, most existing forecasting models are region-specific and depend on large datasets, limiting their applicability across different climates and geographical areas. These models often lack flexibility and may not perform well in regions with limited historical data, leading to inaccurate predictions. This paper proposes a global model for 24-hour-ahead hourly electrical energy demand prediction that is designed to perform effectively across diverse climate conditions and datasets. The model's efficiency is demonstrated using data from two distinct regions: Ireland, with a maritime climate and Vietnam, with a tropical climate. Remarkably, the model achieves high accuracy even with a limited dataset spanning only nine months. Its robustness is further validated across different seasons in Ireland (summer and winter) and Vietnam (dry and wet). The proposed model is evaluated against state-of-the-art machine learning and deep learning methods. Simulation results indicate that the model consistently outperforms benchmark models, showcasing its capability to provide reliable forecasts globally, regardless of varying climatic conditions and data availability. This research underscores the model's potential to enhance the efficiency and sustainability of Energy Communities worldwide. The proposed model achieves a Mean Absolute Percentage Error of 8.0% and 4.0% on the full Irish and Vietnamese datasets.

## 1. INTRODUCTION

The electric demand prediction plays a significant role in the Energy Management System (EMS), which allows an EMS to optimise its objectives for future periods depending on the time horizons required by the EMS such as hour-ahead, day-ahead and week-ahead optimisation so, each optimisation required corresponding predictions of intermittent profiles including Renewable Energy Sources (RESs), electric demand, and electricity prices [1] to minimise or maximise the objective function [2], [3]. The prevailing methods for electric demand prediction in the literature are AutoRegressive Integrated Moving Average (ARIMA), ARIMA with Exogenous Variables (ARIMAX), Support Vector Regression (SVR), eXtreme Gradient Boosting (XGBoost) and Regularisation based Long Short-Term Memory (RLSTM), which have been implemented for different types of electric loads and time horizons. There are three main types of electric loads: residential, commercial and industrial. The focus of this research study is residential electric load.

An Energy Community (EC) is a group of energy providers and users in a specific area working together to meet their energy needs sustainably [4]. This idea has been implemented across Ireland and Europe to promote RES [5], [6] and a smart EMS. These ECs require an accurate prediction profile to optimise their EMS and facilitate the electricity prosumers - reducing their electricity tariff and providing a smooth electricity supply.

In literature, there has been much research in different regions for different time horizons with varying resolutions and data training sets. In [7], a methodology for forecasting UK electricity demand was proposed using a lumped forecasting model with a 24-hour prediction horizon. This model incorporated dimensionality reduction through Piecewise Aggregate Approximation (PAA) and Symbolic Discretisation (SAX). For a 7-day prediction period, the model achieved an average R-squared (R²) value of 0.92, an average Root Mean Square Error (RMSE) of 0.534, an average Mean Absolute Error (MAE) of 7.693 and an average Mean Absolute Percentage Error (MAPE) of 26.169%. Here, the proposed model is benchmarked against the Holt-Winters and Naïve2 models. The Holt-Winters model, evaluated over a 336-hour (two-week) forecast horizon, demonstrates superior performance with an average MAE of 1.380 and an average MAPE of 4.556%, highlighting its effectiveness due to regular updates. The Naïve2 model, also evaluated over a 336-hour horizon, shows a higher average MAE of 11.361 and an average MAPE of 40.247%, indicating that the proposed model performed better than Naïve2 but falls short compared to Holt-Winters. Nguyen et al. [8] proposed a short-term load forecasting model using a Back-propagation Neural Network (BPNN) trained on data from the ISO New England electricity market from 2019 to 2021. The model incorporated features like temperature, humidity, past load and real-time price (RTP). The inclusion of RTP significantly improved the model's accuracy, reducing the MAPE from 2.08% to 1.44% for 24-hour ahead forecasts. Also, the model demonstrated promising results for one-week and one-month ahead forecasts, with MAPE values of 2.1% and 3.03%, respectively.

*This research work is supported by Science Foundation Ireland Centre for Research Training focused on Future Networks and the Internet of Things (AdvanceCRT), under Grant number 18/CRT/6222.*
*[1]Dept. of Elect. & Electronic Eng., Munster Technological University, Cork, T12 P928, Ireland.*
*[2]Nimbus Research Centre, Munster Technological University, Cork, T12 P928, Ireland.*
*[3]Ho Chi Minh City University of Technology, Vietnam National University Ho Chi Minh City, Ho Chi Minh, Vietnam.*
*[*]Corresponding author:* Phan Quang An; Email: pqan@hcmut.edu.vn.

Researchers in [9] compared seven data-driven models for city-scale daily electricity usage prediction, including a five-parameter change-point model, a Heating/Cooling Degree Hour model, a time series decomposed model (Facebook Prophet) and Machine Learning (ML) models (Random Forest (RF), Support Vector Machine (SVM), Neural Network and Gradient Boosting Machine). The models were evaluated using hourly electricity usage data from three US metropolitan areas - Los Angeles, Sacramento and New York - spanning from July 2015 to September 2020. The Gradient Boosting Machine (lightGBM) achieved the best performance, with a Coefficient of Variation of the RMSE (CVRMSE) of 6.5% for Los Angeles, 4.6% for Sacramento and 4.1% for New York. Metrics such as MAE, RMSE and CVRMSE were employed to assess model accuracy, highlighting lightGBM's superior predictive accuracy in comparison to other methods.

In [10], an Artificial Neural Network (ANN) model, optimised using Particle Swarm Optimization algorithms, was employed to forecast long-term Electrical Energy Consumption (EEC) for the United States, Organisation for Economic Co-operation and Development (OECD), China, India and Iran. The dataset spans from 1990 to 2019, with forecasts extending from 2020 to 2050. The model was evaluated using MAE, MAPE, RMSE and MSE. The best MAPE values achieved with the E-PSO-ANN model were 1.01% for the USA, 1.36% for the OECD, 1.27% for China, 1.04% for India and 1.53% for Iran. Rueda et al.

[11] investigated a WaveNet-based encoder-decoder architecture for 24-hour-ahead load demand forecasting in France, using Réseau de Transport d'Électricité (RTE) data from 2017-2019. WaveNet, utilising dilated causal convolutions, outperformed models like ARIMA, LSTM and Gated Recurrent Unit (GRU). The model achieved the lowest RMSE of 1638.709 MW and MAPE of 2.626% during testing. The proposed approach forecasts at a daily time scale, demonstrating superior accuracy and robustness across different seasons compared to traditional methods.

The study by authors [12], evaluated various electric load forecasting models, applying methods from classic ML, Deep Learning (DL) and dynamic mode decomposition to a dataset of real electricity load from a Spanish utility over three years. Models were assessed for various forecast horizons (day/week/month) using metrics like MSE and Symmetric MAPE (sMAPE). The top-performing model was a DL architecture with One-Dimensional Convolutional Neural Network (1D-CNN) and LSTM layers, achieving an R² of 0.9087, an sMAPE of 5.6934 and a Relative RMSE (RRMSE) of 0.0797 for 1-month forecasts.

The study examined electricity load forecasting for Bali Island, Indonesia, using data from 2018–2019 [13]. Machine learning models, specifically Generalised Regression Neural Network (GRNN) and SVM, were applied to forecast electricity load based on weather parameters. The GRNN model achieved the best performance with a Correlation Coefficient (CC) of 0.937 and a RMSE of 41.72, while the SVM model had a CC of 0.934 and an RMSE of 48.88. Forecasts were made for a 1-month horizon using historical data and the models incorporated hourly and daily electricity consumption patterns. In [14], the authors conducted a study focusing on short-term electricity load forecasting using LSTM networks with stacking and time-step techniques. The simulations were done on hourly transmission data from Indonesia spanning 2013–2017, the best model achieved a MAPE of 8.63 with a three-layer LSTM network and 1024 nodes. The model forecasted the day-ahead power based on data from the previous three periods, demonstrating improved accuracy compared to other models.

Zaim et al. [15] proposed a forecasting model for weekly maximum electricity demand in Malaysia, combining Seasonal ARIMA (SARIMA) with Generalised Autoregressive Conditional Heteroskedasticity (GARCH). The study used weekly data spanning from 2005 to 2016 and demonstrated that the SARIMA (1,1,0) (0,1,0) 52-GARCH (1,2) model with Generalised Error Distribution (GED) achieved the best performance. This model reached a MAPE of 3.13%, indicating high forecasting accuracy. The one-step-ahead forecasts effectively account for seasonal volatility and outperformed other models in accuracy. Songkin et al. [16] evaluated short-term load forecasting using datasets from Sabah Grid (2010-2018) and Spanish Grid (2015-2019). Various models, including Exponential Smoothing, Moving Average, ARIMA, machine learning methods like Multilayer Perceptron (MLP), Decision Tree Regression (DTR), SVR and Recurrent Neural Networks (RNNs), were assessed. The ensemble model combining MLP, DTR and Gradient Boosting achieved the best performance, with MAPE of 0.83% for Sabah and 4.47% for Spain. The study highlighted the ensemble model's superior accuracy in handling non-linear and complex datasets over traditional and individual ML methods.

In [17], a model (Stepwise Regression-Variational Mode Decomposition-Empirical Mode Decomposition-Long Short-Term Memory-Grid Search Optimisation (SR-VMD-EMD-LSTM-GSO)) was proposed for predicting daily electricity peak demand in Thailand during the COVID-19 pandemic, combining Stepwise Regression (SR), Variational Mode Decomposition (VMD), Empirical Mode Decomposition (EMD), LSTM networks and Grid Search Optimisation (GSO). The model was tested on data from January 2018 to December 2020, with a one-day-ahead forecast time scale. It was compared against benchmarks including ARIMA, Holt-Winters and single LSTM models, showing superior performance with a MAPE of 3.07%, RMSE of 999 and MAE of 762, demonstrating its robustness in handling pandemic-induced disruptions.

The authors in [18], proposed a day-ahead load forecasting approach using smart meter data from Irish residential customers between July 2009 and December 2010. The method involved clustering customers using K-means based on daily load fluctuations, followed by forecasting each cluster using a Non-linear Autoregressive Neural Network (NAR) and aggregating the results. The model achieved its best performance with 35 clusters, yielding a MAPE of 4.27% and an RMSE of 249.32kW, outperforming two benchmark methods: Method A, which aggregated all customers without clustering and Method B, which employed clustering but differs in approach. The proposed method consistently demonstrates enhanced accuracy at the specified aggregation level. In [19], two forecasting methods - ARIMA and ANN were evaluated for predicting daily electricity demand using data from 709 Irish households over 12,865 hours from 2009 to 2010. The time scale of the data was hourly, with forecasts based on daily aggregated load values. The ANN model outperformed ARIMA, achieving a MAPE of 1.80% compared to ARIMA's 2.61%. Forecasting was conducted for an 18-month period, with ANN demonstrating superior accuracy in handling non-linear data. The study underscored the ANN's effectiveness in capturing complex demand patterns, suggesting its advantage over traditional methods for future load prediction.

In [20], the authors proposed a method to forecast electricity consumption using SVR, MLP and linear regression on data from 782 Irish homes. For district-level predictions, SVR achieves the best results, with 1-hour and 24-hour ahead MAPE of 3.4% and 4.3%, respectively, outperforming other methods. However, for individual households, linear regression proved more accurate, with 1-hour and 24-hour NRMSE of 0.56 and 0.61, respectively. The authors also found that clustering houses improves SVR predictions slightly, but temperature data does not enhance accuracy. Ngo et al. [21], proposed a hybrid AI model Seasonal Autoregressive Integrated Moving Average and Firefly-inspired Optimised (SAMFOR) was proposed for 24-hour-ahead energy consumption forecasting in buildings using a dataset from Danang, Vietnam, collected over 2018-2019 at 30-minute intervals. The model combined SARIMA, SVR and Firefly-inspired optimisation to enhance prediction accuracy. Sensitivity analyses determined optimal inputs, lag values and training data length. The SAMFOR model outperformed other methods with an RMSE of 1.77kWh, MAPE of 9.56% and an R-value of 0.914. The study highlighted the model's effectiveness for short-term energy forecasting at a 30-minute time scale.

Based on the findings from the literature review, this research focuses on developing a day-ahead prediction model that can adapt to a wide range of weather conditions. Unlike previous studies, this research enhances the diversity of input data by incorporating datasets from multiple climatic regions. Furthermore, all the data used in this study are real-world observations, ensuring the model's parameters are highly applicable in practical scenarios. The key contribution of this research is that the developed model's results can be utilised to optimise the operation of Community Microgrids (CMGs) across different geographical locations, enhancing energy management and efficiency in various environments [22], [23].

This paper is structured as follows: Section 2 describes data collection and cleaning, whilst Section 3 explains the methodology of proposed and benchmark models. Section 4 evaluates the performance of the proposed XGBoost model. Finally, Section 5 details principal conclusions and future work.

## 2. DATA DESCRIPTION

In this study, electric demand and temperature data from two distinct geographical locations, Ireland and Vietnam, are utilised to assess the effectiveness of the proposed XGBoost model across varied climatic conditions. The selection of these datasets is strategic and aimed at exploring the model's adaptability and accuracy in predicting electricity demand under different environmental influences. Figures 1 and 2 present the full datasets, visually capturing the underlying patterns and variations. While a slight trend is noticeable, significant fluctuations in both Irish and Vietnamese datasets introduce complexity, posing challenges for predictive models.

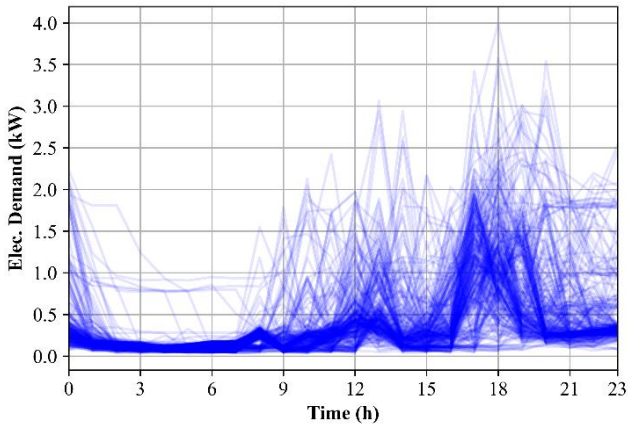

**Fig. 1. Irish residential daily load from 23rd September 2023 to 6th July 2024.**

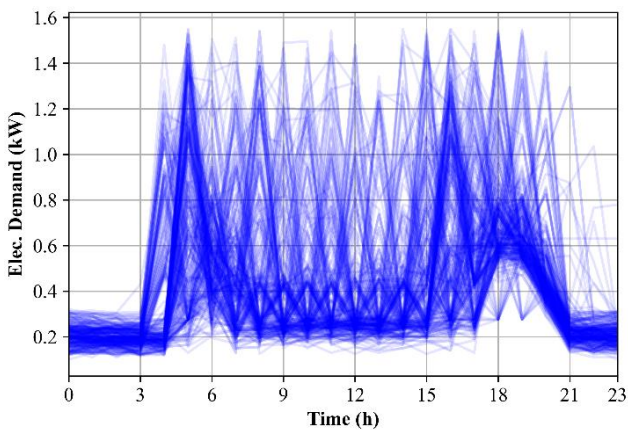

**Fig. 2. Vietnamese residential daily load from 23rd September 2023 to 6th July 2024.**

### 2.1. Irish Residential Load

The dataset from Ireland was acquired from a residential household within an EC located in Dublin. Data was collected at a fine temporal resolution, with measurements recorded at one-minute intervals. The dataset includes two primary variables: ambient air temperature and electricity demand.

### 2.2. Vietnamese Residential Load

The Vietnamese dataset was also sourced from a residential household within an EC in Ho Chi Minh, Vietnam. It was recorded at 30-minute intervals, capturing both ambient temperature and electricity demand. The decision to focus on these variables was driven by their direct impact on electricity consumption patterns, particularly in a region with distinct wet and dry seasons.

## 3. METHODOLOGY

This section outlines the six-step process for electrical load prediction, from raw data collection to result export and analysis.

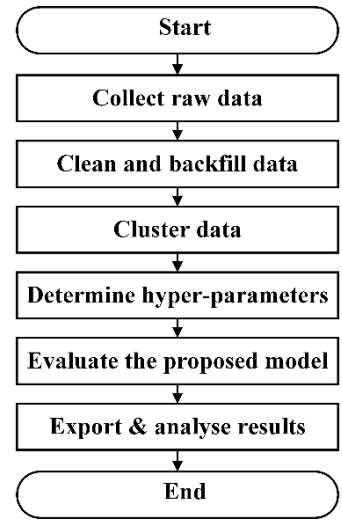

**Fig. 3. Flow Chart of 24-hr-ahead electric Load prediction.**

### 3.1. Forecast Procedure

Figure 3 outlines the complete forecasting model process, starting from the collection of raw data to the exporting and analysis of results in Step 6.

Initially, the Irish raw energy data was recorded in a minute time slot where energy was in (Wh), which had been summed over an hourly time slot. Here the temperature data was averaged over an hour to convert the whole data to an hourly time scale. It was further subjected to a thorough examination for outliers, as shown in Step 2 of Figure 3. Outliers, defined as data points significantly deviating from the expected range, were identified using statistical techniques such as histograms, box plots and z-score analysis. A Z-score threshold was employed to systematically detect these anomalies, which were then addressed to prevent skewing the model's results. The identified outliers were replaced using data imputation techniques that preserved the overall trend and integrity of the time series.

Handling missing values was another essential part of the data preparation process. Unaddressed missing data points can introduce biases and reduce predictive accuracy. Initially, missing values were imputed using data from the previous hour, a method effective for short gaps. However, for longer periods of missing data, data from the same hour of the previous week was used, leveraging the weekly cyclic nature of electricity demand. This approach ensured dataset continuity and preserved the characteristic patterns of electricity consumption. The final Irish dataset, after processing, spanned from 23rd September 2023 to 6th July 2024, covering multiple seasons and providing a robust basis for testing the predictive performance of the XGBoost model. Data was then clustered into summer (1st June 2023 to 31st August 2023) and winter (1st December 2023 to 31st March 2024) seasons to analyse the prediction models for each season, as depicted in Step 3 of Figure 3.

Similarly, the Vietnamese raw data underwent rigorous pre-processing, as shown in Steps 1 to 3 of Figure 3. Initially averaging the 30-minute time-scaled raw data both power and temperature over an hourly time slot to ensure suitability for day-ahead hourly model requirements and synchronisation with Irish data. Outliers were identified in a similar manner with a threshold Z-score of 3 applied to flag significant deviations. These outliers were replaced with the median value of the dataset to minimise the impact on the overall distribution and maintain the dataset's representative nature.

Handling missing values in the Vietnamese dataset required a tailored approach due to unique regional consumption patterns. For isolated missing values, the last valid observation was used as a substitute. For larger gaps, data from the same time of the previous week was employed, like the method used for the Irish dataset, ensuring that imputed values reflected typical consumption patterns while accounting for both daily and weekly cycles in electricity demand. The processed Vietnamese dataset, like the Irish counterpart, spanned from 23$^{rd}$ September 2023 to 6$^{th}$ July 2024. This synchronisation across datasets from different geographical locations not only facilitates comparative analysis but also enables comprehensive evaluation of the forecast model's performance across diverse climatic and cultural contexts. Additionally, the data was grouped into dry (23$^{rd}$ September 2023 to 31$^{st}$ March 2024) and wet (1$^{st}$ April 2024 to 6$^{th}$ July 2024) seasons to explore prediction models for each season, as indicated in Step 3 of Figure 3.

The careful pre-processing and alignment of the Irish and Vietnamese datasets are crucial for the robust evaluation of day-ahead electricity demand forecasting models. In this study data time periods were aligned to ensure consistency which provides a solid foundation for comparing model performance across different environments. The standardised hourly datasets from both Ireland and Vietnam enable detailed investigation into how local conditions such as temperature and seasonal variations affect electricity demand patterns, allowing for a more nuanced understanding of the proposed model's adaptability and accuracy.

Simulations were conducted across various periods and locations, including the full dataset from 23$^{rd}$ September 2023 to 7$^{th}$ July 2024 for both Ireland and Vietnam, the summer and winter seasons in Ireland and the dry and wet seasons in Vietnam. Hyperparameters for each model were optimised using grid search with cross-validation over five folds.

Step 4 focuses on optimising hyperparameters using a grid search approach that evaluates all the possible combinations within a predefined parameter space to identify the optimal configuration. This method exhaustively tests each combination of hyperparameters and selects the one that achieves the best performance based on average MAE across all folds. Step 5 involves evaluating the proposed model's performance on the test set using a suite of metrics: MAE, Mean Squared Error (MSE), RMSE, MAPE and the R² score. Finally, Step 6 is dedicated to exporting all results and analysing them. Steps 4 to 6 are further detailed in the Results and Discussion section.

Overall, the forecast procedure, as outlined in Figure 3, ensures a systematic approach to data collection, pre-processing, model optimisation and evaluation. Meticulously preparing the datasets and carefully tuning the models, this study aims to deliver accurate and reliable electricity demand forecasts across various regions and seasons.

### 3.2. Prediction Algorithms

This section describes the proposed and benchmarked prediction methods, including both ML and DL approaches.

#### 3.2.1. eXtreme Gradient Boosting (XGBoost)

The XGBoost algorithm, proposed by Tianqi Chen in 2016 [24], is a decision tree-based model that employs a boosting technique to enhance predictive accuracy. Boosting functions by sequentially training a series of weak tree models, each of which compensates for the residual errors of the previous model. However, despite its ability to improve accuracy, boosting can be computationally intensive and prone to overfitting. XGBoost mitigates these drawbacks by incorporating tree pruning, parallelisation and regularisation terms, which collectively improve the efficiency and robustness of the boosting process. Although XGBoost has recently gained traction across various fields, its application in load forecasting remains relatively underexplored [25]. In this research, XGBoost is applied for day-ahead load forecasting, leveraging its various advantages. The predicted value using the XGBoost regressor is computed as shown in (3.1) [26].

$$\hat{y}_t = \sum_{k=1}^{K} f_k(x_t), f_k \in \mathfrak{f} \tag{3.1}$$

Where, $\hat{y}_t$ denotes the predicted value, $f_k(.)$ represents the kth tree model, $x_t$ indicates the input feature, K is the number of trees and $\mathfrak{f}$ denotes the functional space containing the set of trees.

$$F_{obj} = \sum_{t=1}^{N} L(y_t, \hat{y}_t) + \sum_{t=1}^{N} \Omega(f_k) \tag{3.2}$$

The objective function in the XGBoost regressor includes a regularisation term and is defined by (3.2) [26]. Here, $L(.)$ represents the loss function, specifically the MAE, $y_t$ denotes the actual value and $\Omega(.)$ represents the regularisation term, which imposes a penalty on model complexity. The regularisation term is further elaborated in (3.3) [26], where $T$ denotes the number of leaves, $\omega_j$ denotes the $j$th vector of scores on leaves and $\lambda$ and $\gamma$ denote the penalty factors.

$$\Omega(f) = \gamma T + \frac{1}{2}\lambda \sum_{j-1}^{T} \omega_j^2 \qquad (3.3)$$

The XGBoost algorithm is a tree-based model that segments data according to input features, using these segmented branches for prediction. To explore it, this study introduces season-specific models: summer, winter and a combined season model for Ireland, as well as dry, wet and a combined season model for Vietnam. These models classify the training data based on the forecast target date to examine the effects of different seasons and locations on the proposed XGBoost model.

### 3.2.2. AutoRegressive Integrated Moving Average (ARIMA)

It is a statistical model used for time series forecasting. It combines three components: Autoregression (AR), Differencing (I) to make the series stationary and a Moving Average (MA) of past errors. It predicts future values based on a linear combination of past observations and errors, making it effective for short-term forecasting when data exhibits patterns such as trends or seasonality.

### 3.2.3. AutoRegressive Integrated Moving Average with Exogenous Variables (ARIMAX)

ARIMAX extends ARIMA by incorporating external variables (exogenous variables) into the model. In this study, temperature serves as an exogenous variable. Besides using past values and errors for prediction, ARIMAX utilises these external factors to enhance forecast accuracy when relevant external information is available. Both ARIMA and ARIMAX were applied in a walk-forward mode, where the predicted energy demand was used iteratively as a feature to improve model accuracy.

### 3.2.4. Support Vector Regression (SVR)

SVR is a type of regression model derived from SVM. It works by finding a hyperplane that best fits the data within a specified margin of tolerance, allowing for some deviation. The goal is to minimise the error within this margin while maximising the margin itself. SVR is particularly effective for capturing complex relationships in the data and is robust to outliers, making it useful for regression tasks where the relationship between variables is non-linear or when dealing with high-dimensional data.

### 3.2.5. Regularisation-based Long Short-Term Memory (RLSTM)

LSTM is a type of RNN designed to address the limitations of traditional RNNs in capturing long-term dependencies. LSTM networks incorporate special gating mechanisms, including the input gate, forget gate and output gate, which regulate the flow of information and help retain relevant data over long sequences. RLSTM incorporates Dropout regularisation into an LSTM architecture to avoid overfitting. This architecture allows RLSTMs to effectively learn and remember information from extended time series, making them well-suited for tasks involving sequential data, such as time series forecasting and natural language processing.

## 4. RESULTS AND DISCUSSION

This section presents the simulation results to demonstrate the effectiveness of the proposed XGBoost model and its configuration. The proposed electric demand forecasting model is evaluated against several benchmark models, including ARIMA, ARIMAX, SVR and RLSTM networks. These models are assessed to identify the most suitable forecasting technique for EC, which operates under highly variable weather conditions and limited datasets.

Simulations are conducted across various periods and locations, including the full dataset from September 23[rd] 2023 to July 7[th] 2024 for both Ireland and Vietnam, the summer season in Ireland, the winter season in Ireland, the dry season in Vietnam and the wet season in Vietnam. Hyperparameters for each model are optimised using grid search with cross-validation over five folds.

Grid search is a methodical approach for tuning hyperparameters that evaluates all possible combinations within a predefined parameter space to identify the optimal configuration. It exhaustively tests each combination of hyperparameters and selects the one that achieves the best performance based on an average MAE on all the folds. For the XGBoost model, the parameter space includes several key hyperparameters. The number of trees in the ensemble denoted as $\eta_{estimators}$, is tested with values [50, 100, 150, 200], controlling the total number of boosting iterations. The *learning rate* is explored with values [0.01, 0.05, 0.1, 0.2], determining the step size at each iteration. The maximum depth of the trees, denoted as *max depth*, is varied with values [3, 5, 7, 10], which limits the complexity of individual trees. The minimum child weight, denoted as *min child weight*, is assessed with values [1, 3, 5, 7], specifying the minimum sum of instance weights required in a child node. The subsampling rate, denoted as subsample, is examined with values [0.7, 0.8, 0.9, 1.0], determining the fraction of observations used for each tree. The column subsampling rate by tree, denoted as *colsample bytree*, is evaluated with values [0.7, 0.8, 0.9, 1.0], indicating the fraction of features used for each tree.

**Table 1. Best hyper-params of the proposed 24-hour-ahead prediction models**

| Hyper-params | Irish Summer | Irish Winter | Vietnam Dry | Vietnam Wet | Irish Full Data | Vietnam Full Data |
|---|---|---|---|---|---|---|
| $\alpha$ | 0 | 0 | 0 | 0 | 0 | 0 |
| colsample bytree | 1.0 | 1.0 | 1.0 | 1.0 | 1.0 | 1.0 |
| $\lambda$ | 0.1 | 0.1 | 0.1 | 0 | 1.0 | 0.1 |
| learning rate | 0.1 | 0.1 | 0.05 | 0.05 | 0.05 | 0.05 |
| max depth | 7 | 10 | 10 | 10 | 10 | 10 |
| min child weight | 5 | 7 | 7 | 7 | 7 | 7 |
| $n_{estimators}$ | 200 | 200 | 200 | 200 | 200 | 200 |
| subsample | 0.7 | 0.7 | 0.7 | 0.7 | 0.7 | 0.7 |

The L2 regularisation term, denoted as $\lambda$, is tested with values [0, 0.1, 1, 10], adding a penalty to prevent overfitting by shrinking weights. The L1 regularisation term, denoted as $\alpha$, is assessed with values [0, 0.1, 1, 10], encouraging sparsity in the model by driving some feature weights to zero. Table 1 presents the optimal hyperparameters for each model.

The datasets undergo a feature engineering process where lagged variables are created to capture temporal dependencies and features are scaled using Standard-Scaler to ensure that all features contribute equally during training. A train-test split is performed, with 80% of the data used for training and 20% for testing.

Time-series-split, a cross-validation technique tailored for time-dependent data, is employed for model training. This technique maintains the temporal order of observations by dividing the training data into multiple sequential folds. Each fold consists of an expanding training set and a corresponding validation set. For instance, in the provided implementation, the training data was divided into five folds, each comprising unique combinations of training and validation subsets. This approach ensures that predictions are based on past observations and preserves the temporal integrity of the model evaluation.

The model is further refined using early stopping, a technique that halts training when performance on the validation set no longer improves, thereby preventing overfitting. Finally, the model's performance is evaluated on the test set using a suite of metrics mentioned earlier.

In the Irish dataset's summer subset (Figure 4, Table 2), the model demonstrated solid performance with an MAE of 0.0384kW and an RMSE of 0.0731kW. The MAPE is equal to 12.0%, indicating that the model's predictions were within 12.0% of the actual values on average during the summer months. Despite this, the R² score remained high at 0.96, showing that the model captured a significant portion of the variance in power demand during this season. When compared with the winter subset, the summer data suggests that the model's performance remains relatively consistent, though some variations in accuracy are present. It can be seen in Table 3 that ARIMA and ARIMAX are inaccurate because they could not capture the non-linearity of electrical load. SVR and RLSTM have better results and it is also necessary to mention that RLSTM has lower accuracy than the proposed model and SVR because DL models perform better on large datasets. The winter subset of the Irish dataset (Figure 5, Table 3) exhibited an MAE of 0.0398kW and an RMSE of 0.0884kW, with a MAPE of 11.0%, slightly improving over the summer subset. The R² score of 0.95, while slightly lower, indicates that the model was still effective in capturing the variance in power demand during winter. This seasonal comparison highlights that while the model performs well across different seasons, the winter conditions in Ireland, characterised by increased variability in power demand, present a more challenging scenario. Similarly, as in summer ARIMA and ARIMAX suffer SVR and RLSTM have better accuracy comparatively and RLSTM struggles with small datasets because of its inherent nature to perform better on large datasets.

When applied to the Vietnam dataset during the dry season (Figure 6, Table 4), the model achieved an MAE of 0.0197kW and an RMSE of 0.0327kW, with a MAPE of 6.0%. The R² score was exceptionally high at 0.99, indicating that the model effectively captured almost all the variance in power demand during the dry season. Compared to the Irish dataset, the results from Vietnam show that the model performs exceptionally well in more stable climate conditions like the dry season, where the variability in power demand is less pronounced. Likewise, as in previous cases, the proposed model has better accuracy than benchmark models. Here, the RLSTM model more inferior accuracy than the Irish summer and winter seasons models because of two reasons: the small dataset and the Vietnamese data are not very complex, so DL models are good at non-linear and large datasets.

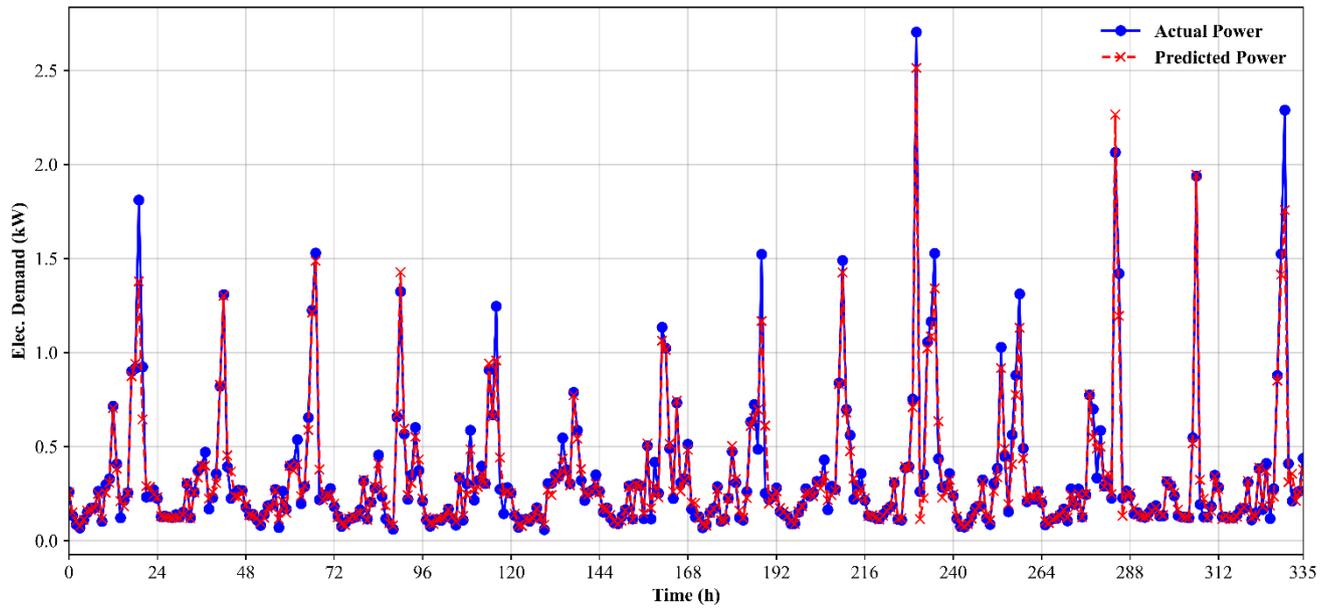

**Fig. 4. 24-hr-ahead electrical load prediction for the Irish summer season (2 weeks).**

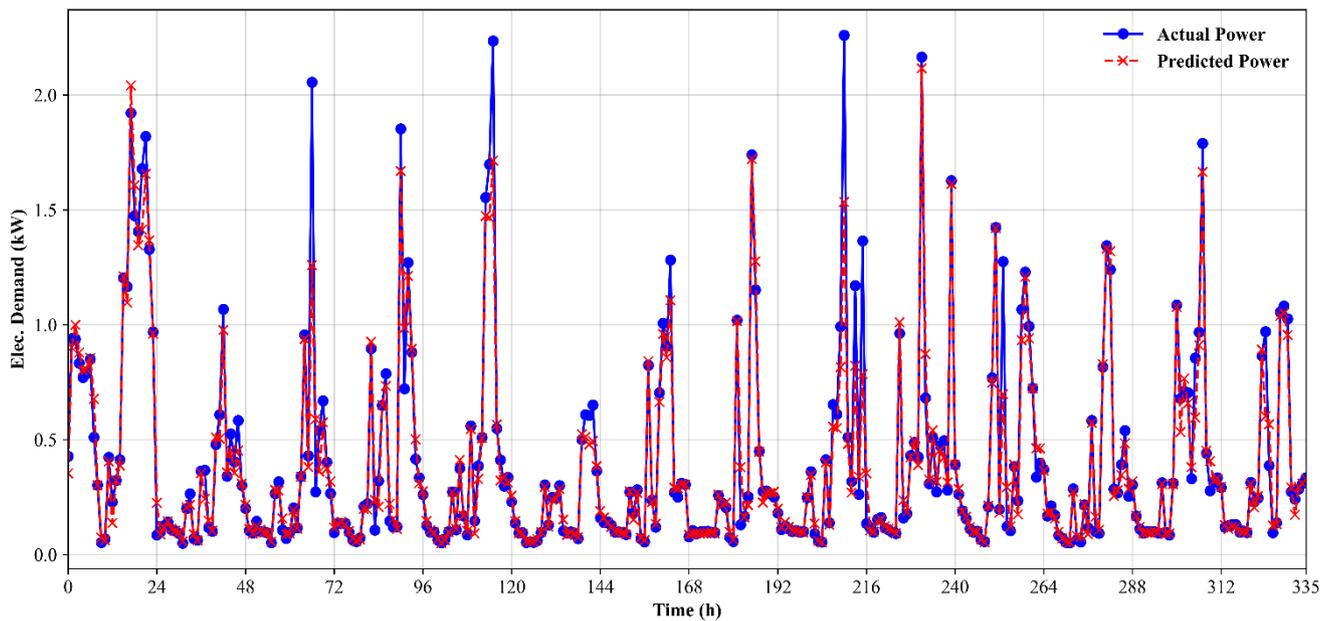

**Fig. 5. 24-hr-ahead electrical load prediction for the Irish winter season (2 weeks).**

**Table 2. Day-Ahead Elec. Load Forecasting Performance Summer Irish data**

| Eval. Metrics | XGBoost | ARIMA | ARIMAX | SVR | RLSTM |
|---|---|---|---|---|---|
| MAE | 0.0384 | 0.2026 | 0.1981 | 0.0545 | 0.0559 |
| MSE | 0.0053 | 0.1018 | 0.1082 | 0.0069 | 0.0060 |
| RMSE | 0.0731 | 0.3289 | 0.3289 | 0.0829 | 0.0774 |
| MAPE | 12.0% | 78.7% | 74.80% | 23.96% | 25.84% |
| $R^2$ | 0.96 | 0.17 | 0.17 | 0.95 | 0.95 |

**Table 3. Day-Ahead Elec. Load Forecasting Performance Winter Irish data**

| Eval. Metrics | XGBoost | ARIMA | ARIMAX | SVR | RLSTM |
|---|---|---|---|---|---|
| MAE | 0.0398 | 0.2242 | 0.2242 | 0.0575 | 0.0438 |
| MSE | 0.0078 | 0.1260 | 0.1267 | 0.0070 | 0.0034 |
| RMSE | 0.0884 | 0.3550 | 0.3559 | 0.0836 | 0.0581 |
| MAPE | 11.0 % | 90.0% | 90.95% | 27.55% | 23.94% |
| $R^2$ | 0.95 | 0.24 | 0.24 | 0.96 | 0.98 |

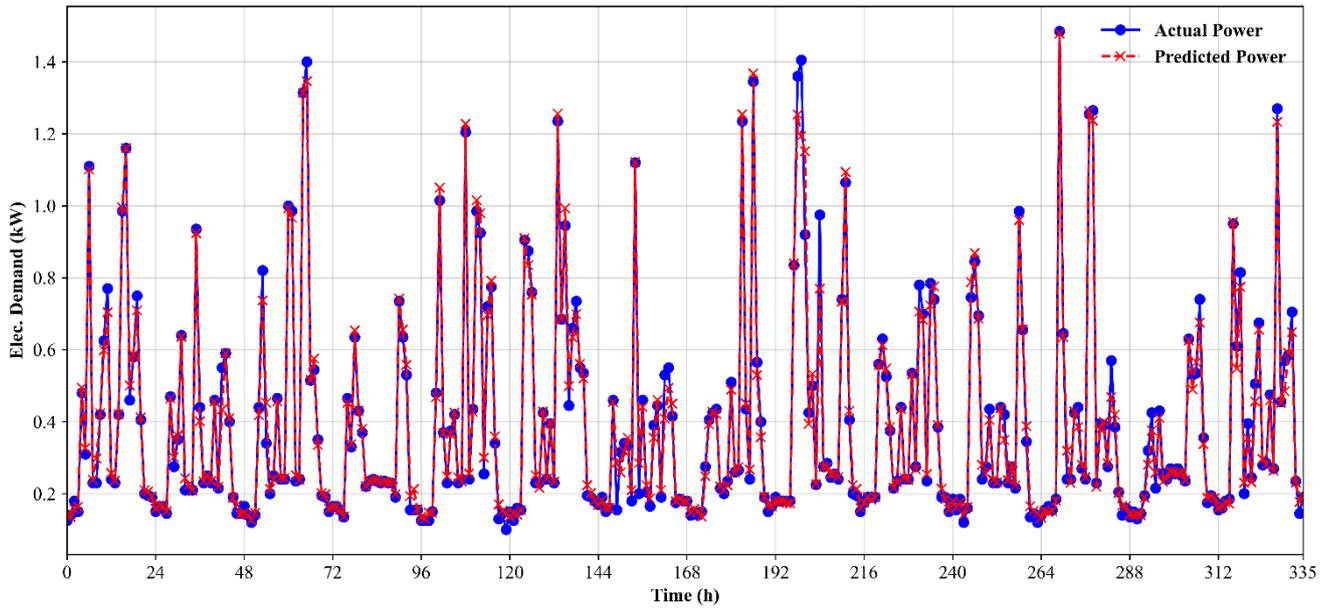

**Fig. 6. 24-hr-ahead electrical load prediction for the Vietnamese dry season (2 weeks).**

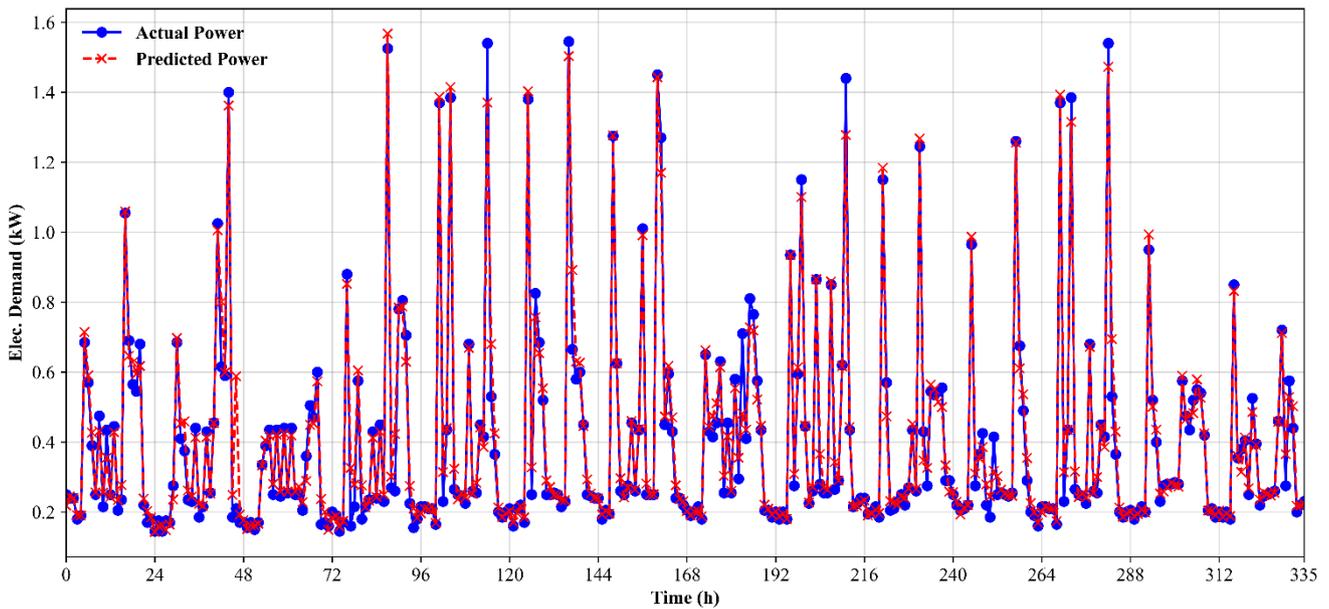

**Fig. 7. 24-hr-ahead electrical load prediction for the Vietnamese wet season (2 weeks).**

**Table 4. Day-Ahead Elec. Load Forecasting Performance Dry Vietnam data**

| Eval. Metrics | XGBoost | ARIMA | ARIMAX | SVR | RLSTM |
|---|---|---|---|---|---|
| MAE | 0.0197 | 0.2039 | 0.2097 | 0.0532 | 0.0792 |
| MSE | 0.0011 | 0.0716 | 0.0759 | 0.0043 | 0.0083 |
| RMSE | 0.0327 | 0.2675 | 0.2755 | 0.0655 | 0.0914 |
| MAPE | 6.0 % | 73.88% | 71.06% | 21.41% | 35.80% |
| $R^2$ | 0.99 | 0.12 | 0.07 | 0.95 | 0.90 |

**Table 5. Day-Ahead Elec. Load Forecasting Performance Wet Vietnam data**

| Eval. Metrics | XGBoost | ARIMA | ARIMAX | SVR | RLSTM |
|---|---|---|---|---|---|
| MAE | 0.0297 | 0.1778 | 0.2029 | 0.0596 | 0.0844 |
| MSE | 0.0027 | 0.0706 | 0.0849 | 0.0067 | 0.0124 |
| RMSE | 0.0522 | 0.2658 | 0.2913 | 0.0818 | 0.1113 |
| MAPE | 8.0 % | 48.72% | 57.01% | 20.34% | 26.23% |
| $R^2$ | 0.97 | 0.1987 | 0.04 | 0.92 | 0.86 |

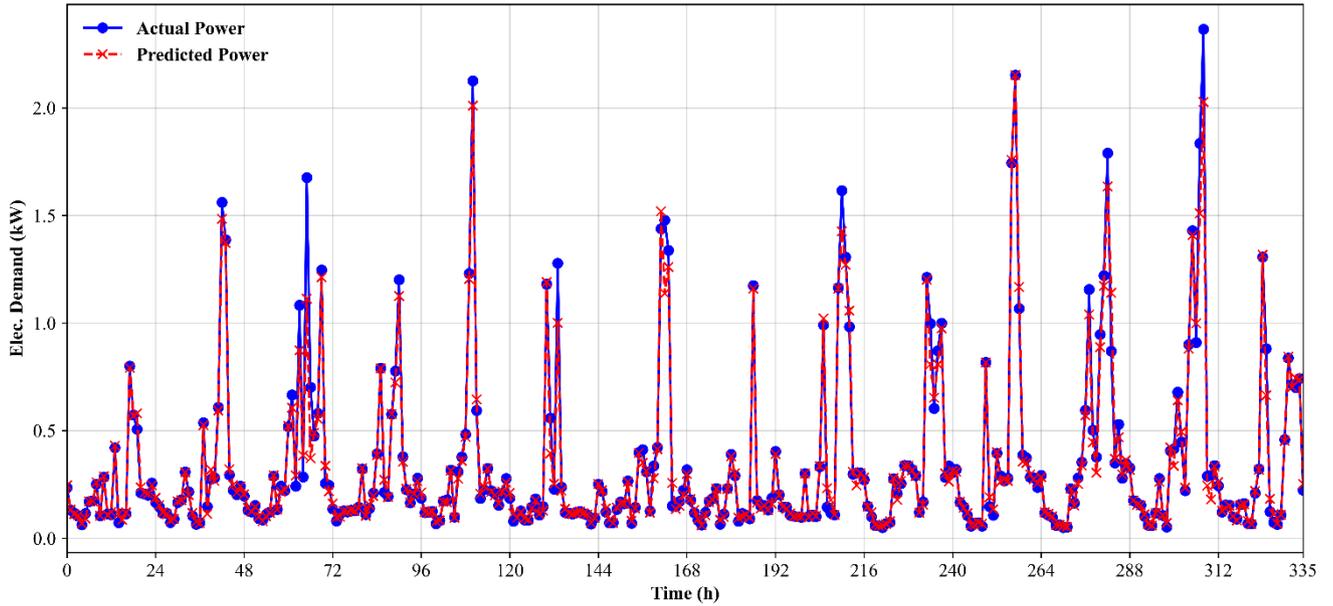

**Fig. 8. 24-hr-ahead electrical load prediction for the Irish full dataset (2 weeks).**

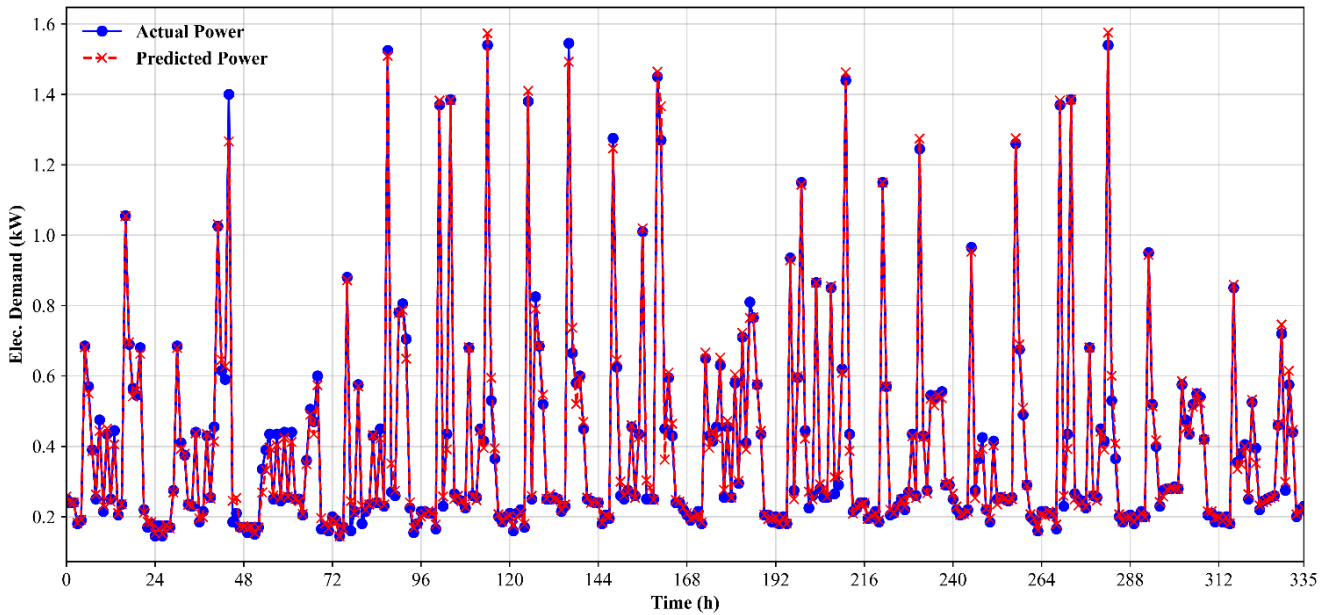

**Fig. 9. 24-hr-ahead electrical load prediction for the Vietnamese full dataset (2 weeks).**

**Table 6. Day-Ahead Elec. Load Forecasting Performance Full Irish data**

| Eval. Metrics | XGBoost | ARIMA | ARIMAX | SVR | RLSTM |
|---|---|---|---|---|---|
| MAE | 0.0276 | 0.2164 | 0.2158 | 0.0850 | 0.0483 |
| MSE | 0.0051 | 0.1312 | 0.1307 | 0.0251 | 0.0056 |
| RMSE | 0.0712 | 0.3622 | 0.3615 | 0.1585 | 0.0751 |
| MAPE | 8.0 % | 86.6% | 86.30% | 42.84% | 20.82% |
| $R^2$ | 0.97 | 0.23 | 0.23 | 0.85 | 0.97 |

**Table 7. Day-Ahead Elec. Load Forecasting Performance Full Vietnam data**

| Eval. Metrics | XGBoost | ARIMA | ARIMAX | SVR | RLSTM |
|---|---|---|---|---|---|
| MAE | 0.0148 | 0.1749 | 0.2014 | 0.0391 | 0.0364 |
| MSE | 0.0007 | 0.0613 | 0.0739 | 0.0026 | 0.0023 |
| RMSE | 0.0265 | 0.2476 | 0.2718 | 0.0511 | 0.0482 |
| MAPE | 4.0 % | 52.93% | 60.75% | 12.23% | 10.99% |
| $R^2$ | 0.99 | 0.20 | 0.04 | 0.97 | 0.97 |

In the wet season of the Vietnam dataset (Figure 7, Table 5), the model's performance showed a slight reduction in accuracy, with an MAE of 0.0297 kW and an RMSE of 0.0522 kW. The MAPE increased to 8.0% and the R² score remained strong at 0.97. The wet season results, when compared to the dry season, suggest that the model is sensitive to the increased complexity of power demand patterns during periods of higher humidity and precipitation. This behavior is similar to the performance difference observed between the Irish summer and winter datasets, where seasonal changes also led to fluctuations in accuracy. In wet season RLSTM has better accuracy than dry season and it is also supporting the above claimed argument that DL models perform better at more complex datasets and wet season has more non-linearity than dry season.

For the full Irish dataset (Figure 8, Table 6), the model achieved an MAE of 0.0276kW and an RMSE of 0.0712kW, with a MAPE of 8.0%. The R² score of 0.97 confirms the model's ability to capture the overall power demand patterns across different seasons in Ireland, despite the challenges posed by the country's more volatile weather. When considering the full Vietnam dataset (Figure 9, Table 7), the model demonstrated consistently strong performance, with an MAE of 0.0148kW and an RMSE of 0.0265kW. The MAPE was low at 4.0% and the R² score was 0.99, underscoring the model's reliability in predicting power demand across both stable and variable conditions. Compared to the full Irish dataset, the Vietnam results underscore the model's adaptability and accuracy in a more stable climate, contrasting with the variability observed in the Irish seasonal subsets. Here, the RLSTM model is performing better than SVR because of its capability to perform well on large datasets but still, the proposed model has outperformed all the benchmark models.

The results for a single day under each scenario are also illustrated in the corresponding figures (Figure 10 - 15). These plots provide a detailed view of the model's daily performance, allowing for a focused evaluation of accuracy across different datasets and conditions and further reinforcing the model's robustness across diverse climates and regions. Overall, the model demonstrates a strong capability to perform well on datasets from two different locations, Ireland and Vietnam. The slightly lower accuracy in the Irish dataset compared to the Vietnam dataset can be attributed to the more volatile nature of weather in Ireland, which presents a greater challenge for power demand prediction. In contrast, the more stable weather patterns in Vietnam allow the model to achieve higher accuracy, highlighting its effectiveness in both stable and variable environments.

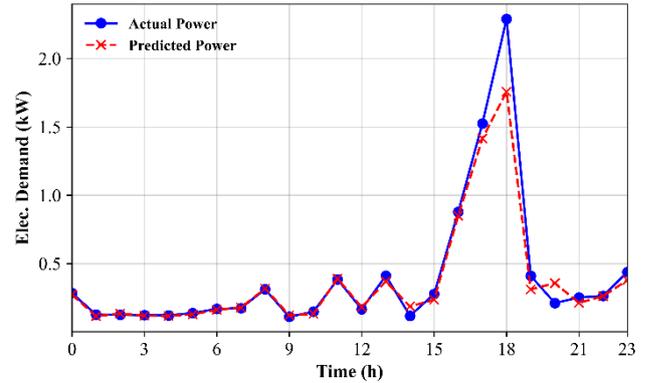

**Fig. 10. Load prediction for the Irish summer season-1 day.**

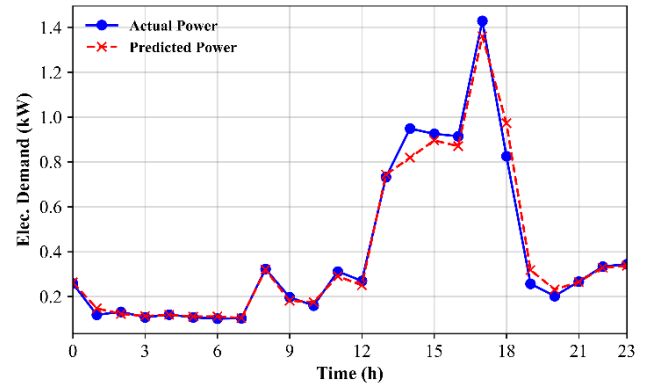

**Fig. 11. Load prediction for the Irish winter season-1 day.**

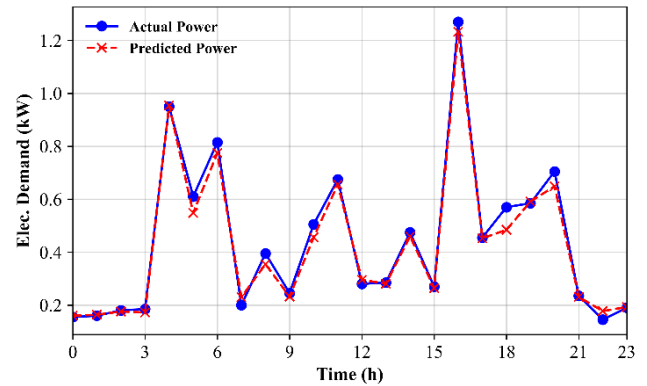

**Fig. 12. Load prediction for the Vietnamese dry season-1 day.**

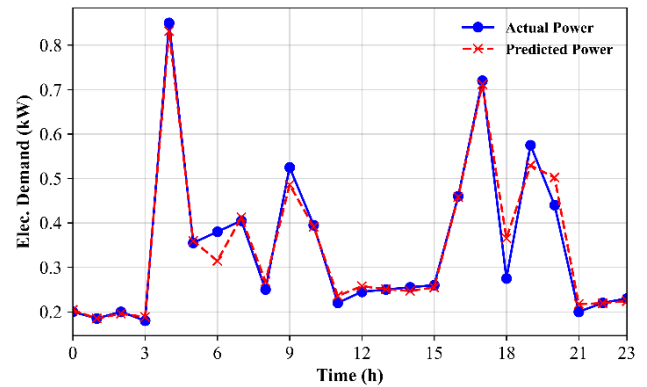

**Fig. 12. Load prediction for the Vietnamese wet season-1 day.**

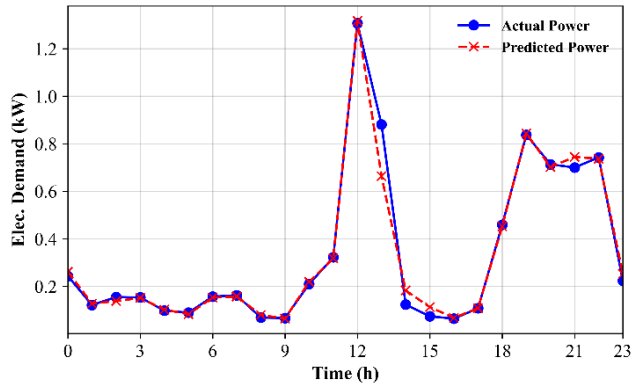

**Fig. 14. Load prediction for the Irish full dataset-1 day.**

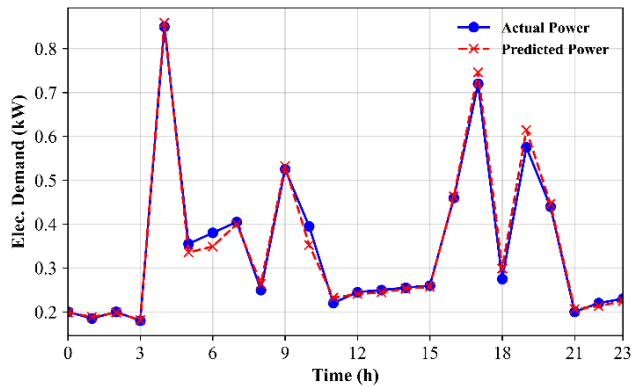

**Fig. 15. Load prediction for the Vietnamese full dataset-1 day.**

## 5. CONCLUSIONS

This research paper presents an XGBoost-based machine learning approach for forecasting 24-hour-ahead hourly electricity demand specifically in houses within a sustainable Energy Community. The model was trained and tested on diverse datasets from different locations in Ireland and Vietnam and its performance was further analysed across different seasons: summer and winter in Ireland and dry and wet seasons in Vietnam. The results indicate that the model achieves a MAPE of 12.0% with an $R^2$ of 0.96 for the summer season in Ireland and a MAPE of 11.0% with an $R^2$ of 0.95 for the winter season. For Vietnam, the model achieved a MAPE of 6.0% with an $R^2$ of 0.99 during the dry season and a MAPE of 8.0% with an $R^2$ of 0.97 during the wet season. When trained and tested on the full dataset spanning from 23rd September 2023 to 6th July 2024, the model demonstrated improved performance, with a MAPE of 8.0% and an $R^2$ of 0.97 for Ireland and a MAPE of 4.0% with an $R^2$ of 0.99 for Vietnam.

The better performance on the Vietnamese dataset can be attributed to its relatively moderate climate, which provides more stable weather conditions compared to the more severe climate conditions in Ireland. The lower accuracy on seasonal datasets, particularly in Ireland, is likely due to the limited amount of data available, which reduces the model's ability to effectively capture seasonal variations. In Vietnam, the wet season's lower accuracy compared to the dry season can be explained by the increased weather variability and severity during this period.

The proposed model was also benchmarked against various machine learning and deep learning methods, including ARIMA, ARIMAX, SVR and RLSTM networks. The ARIMA and ARIMAX struggled overall on all the seasons and full datasets. SVR has better accuracy than RLSTM on small datasets which has bad results. RLSTM performs better than SVR on complex and large datasets. Overall, the simulation results demonstrate that the proposed XGBoost model consistently outperforms these benchmark models, showcasing its efficacy across different climatic conditions and datasets.

For future research, this model will be integrated into a data-driven EMS for EC to explore the impact of energy profiles in managing and optimising energy use within ECs.

## ACKNOWLEDGEMENTS

We acknowledge Ho Chi Minh City University of Technology (HCMUT), VNU-HCM for supporting this study.